\documentclass[journal]{IEEEtran}
\usepackage{amsmath,amsfonts}
\usepackage{algorithm}
\usepackage{array}
\usepackage[caption=false,font=scriptsize,labelfont=sf,textfont=sf]{subfig}
\usepackage{textcomp}
\usepackage{stfloats}
\usepackage{url}
\usepackage{verbatim}
\usepackage{graphicx}
\usepackage{cite}
\usepackage{hyperref}
\usepackage{cleveref}
\usepackage{booktabs}

\usepackage[accsupp]{axessibility}  
\usepackage{multirow}
\usepackage{tabularx} 
\usepackage{color}

\begin{document}
\title{Boosting Salient Object Detection with Knowledge Distillated from Large Foundation Models}

\author{Miaoyang He, Shuyong Gao, Tsui Qin Mok, Weifeng Ge, Wengqiang Zhang,~\IEEEmembership{Member,~IEEE,}
\thanks{This paper was produced by the IEEE Publication Technology Group. They are in Piscataway, NJ.}
\thanks{Manuscript received April 19, 2021; revised August 16, 2021.}}

\markboth{Journal of \LaTeX\ Class Files,~Vol.~14, No.~8, August~2021}%
{Shell \MakeLowercase{\textit{et al.}}: A Sample Article Using IEEEtran.cls for IEEE Journals}

\IEEEpubid{0000--0000/00\$00.00~\copyright~2021 IEEE}

\maketitle

\begin{abstract}
Salient Object Detection (SOD) aims to identify and segment prominent regions within a scene. Traditional models rely on manually annotated pseudo labels with precise pixel-level accuracy, which is time-consuming. We developed a low-cost, high-precision annotation method by leveraging large foundation models to address the challenges. Specifically, we use a weakly supervised approach to guide large models in generating pseudo-labels through textual prompts. Since large models do not effectively focus on the salient regions of images, we manually annotate a subset of text to fine-tune the model. Based on this approach, which enables precise and rapid generation of pseudo-labels, we introduce a new dataset, BDS-TR. Compared to the previous DUTS-TR dataset, BDS-TR is more prominent in scale and encompasses a wider variety of categories and scenes. This expansion will enhance our model's applicability across a broader range of scenarios and provide a more comprehensive foundational dataset for future SOD research. Additionally, we present an edge decoder based on dynamic upsampling, which focuses on object edges while gradually recovering image feature resolution. Comprehensive experiments on five benchmark datasets demonstrate that our method significantly outperforms state-of-the-art approaches and also surpasses several existing fully-supervised SOD methods. The code and results will be made available.

\end{abstract}


\section{Introduction}
\IEEEPARstart{S}{alient} Object Detection (SOD) aims to accurately detect and segment the most attention-grabbing regions within a given image, simulating human attention mechanisms~\cite{gopalakrishnan2009salient,dong2016human,wang2021salient,zhang2022saliency,bhunia2023sketch2saliency}. This technique finds extensive application across various domains of computer vision, including object recognition~\cite{flores2019saliency,lou2023predicting}, image captioning~\cite{zhou2019re}, and person re-identification~\cite{he2020guided}, thereby garnering increasing interest.

\begin{figure*}[!htb]
    \centering
    \includegraphics[width=\textwidth]{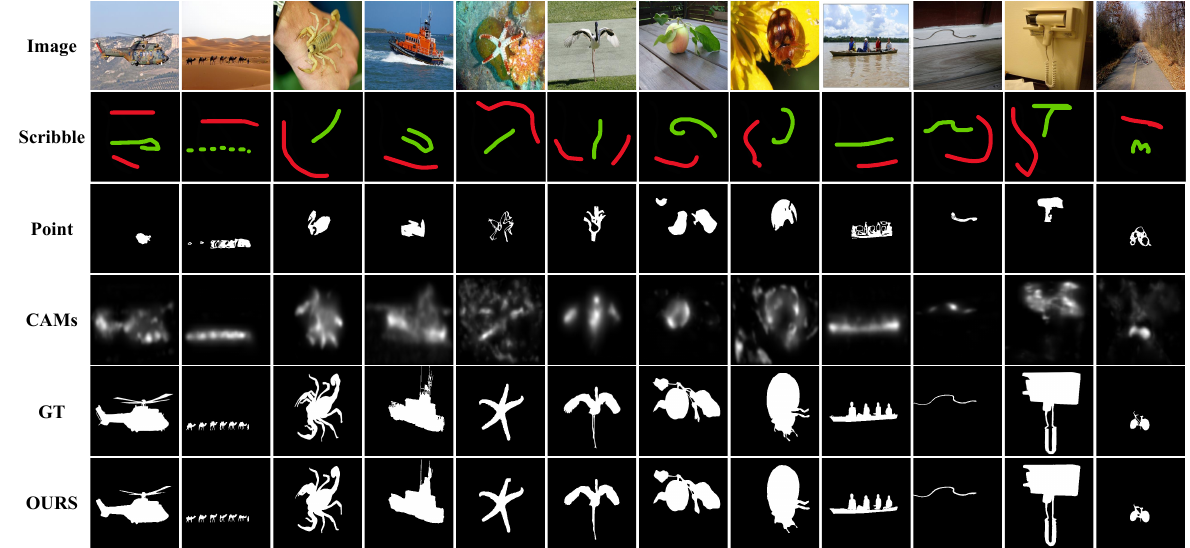}
    \caption{Visual comparison of pseudo-label generation between our method and other approaches. Each row, from top to bottom, shows Image, GT, scribble labels~\cite{zhang2020weakly}, point labels~\cite{gao2022weakly}, CAMs~\cite{zhou2016learning}, and ours. Compared to other weakly supervised methods, our approach generates high-precision pseudo-labels.}
    \label{fig:pseudo}
\end{figure*}

With the breakthroughs in deep learning, state-of-the-art algorithms have achieved remarkable success across various domains. However, currently popular SOD datasets, such as MSRA-B~\cite{li2016deep}, PASCAL-S~\cite{li2014secrets}, ECSSD~\cite{yan2013hierarchical}, and DUTS-TR~\cite{wang2017learning}, rely on precise manual annotation and contain limited sample quantities and categories, thereby constraining the generalization capabilities of existing algorithms in broader application scenarios. Thus, unsupervised or weakly supervised approaches have become more viable alternatives for real-world scenarios, yet these methods are not without challenges. 

Recent unsupervised model~\cite{luo2020n} often acquire saliency cues via traditional SOD methods~\cite{ji2022promoting,wang2022multi,zhang2018deep,cong2022does}, and increasing attention has been directed toward methods using sparse annotations, aiming to achieve a balance between annotation efficiency and model performance. Approaches incorporating image classification labels~\cite{wang2017learning}, scribble annotations~\cite{zhang2020weakly}, and point annotations~\cite{gao2022weakly} offer promising alternatives to reduce manual labeling efforts. Some previous studies have synthesized pseudo-labels from image-level category labels using Class Activation Maps (CAMs)~\cite{zhou2016learning}. Other techniques leverage the rich appearance information in RGB images to refine CAMs~\cite{piao2021mfnet,piao2022noise}. However, in complex scenes, these techniques frequently struggle to capture accurate saliency cues, often deviating from target objects, especially around edges and finer object details, as shown in ~\cref{fig:pseudo}.

This study addresses the above challenges by developing a cost-effective approach to substitute the expensive human labeling process. Our research aims to explore whether the development of large foundation models enables us to boost salient object detection through knowledge distilled from these models. Large foundation models can effectively simulate human perception of complex objects: when accurately identifying the edges and details of objects is challenging, we rely on cognitive understanding to make judgments. Consequently, we propose a framework for generating pseudo-labels using text-guided large models, which offers several advantages. First, compared to labor-intensive manual labeling, this method achieves accurate annotations without prior training, significantly reducing the difficulty and time associated with labeling. Furthermore, textual descriptions can provide large models with additional cues that are not obtainable from images alone, enabling the generation of more precisely detailed labels. As a result, the quality of the generated pseudo-labels has been significantly improved compared to previous methods.
\IEEEpubidadjcol
We utilize the Blip~\cite{li2022blip} model to generate textual descriptions of images. However, Blip cannot focus on the salient parts of images to create appropriate phrases. To address this, we randomly select a small subset of manually annotated text from DUTS-TR to fine-tune Blip, enabling us to generate the necessary textual descriptions for the entire dataset. Then, we employ GroundingDINO~\cite{liu2023grounding} to create bounding boxes for salient objects based on the above textual information. Subsequently, we input the bounding boxes and RGB images into SAM~\cite{kirillov2023segment} to produce the final pseudo-labels. Unlike general object detection, our focus is on identifying the salient objects within a given scene while ignoring non-salient backgrounds and irrelevant objects~\cite{zhang2020weakly,zeng2019multi,song2023towards,yu2021structure,zhuge2022salient}. To better differentiate the saliency of objects, we emphasized the salient characteristics of the objects during the manual text annotation process. This allows BLIP, when fine-tuned, to focus on the salient objects more effectively. Though using SAM for pseudo-label segmentation may not fully align with the traditional concept of weakly supervised SOD, we believe that applying large foundation models has excellent value and potential for future advancements.

\begin{figure*}[!t]
\centering
\subfloat[\fontsize{6}{10}\selectfont Categories distribution in BDS-TR and DUTS-TR.]{\includegraphics[width=0.45\linewidth]{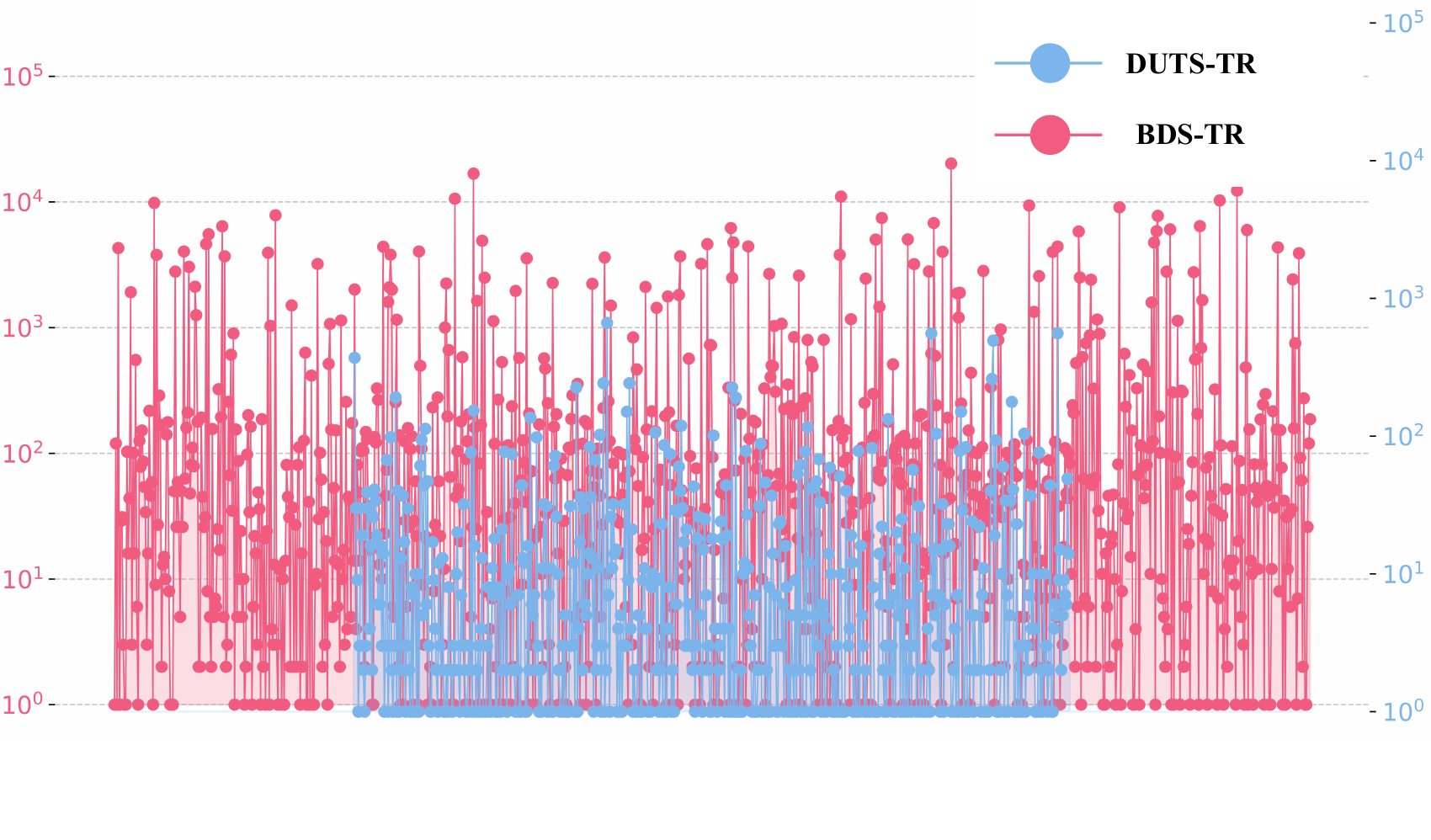}%
\label{fig:line_chart}}
\subfloat[\fontsize{6}{10}\selectfont Subcategory Distribution of Transportation in BDS-TR and DUTS-TR]{\includegraphics[width=0.45\linewidth]{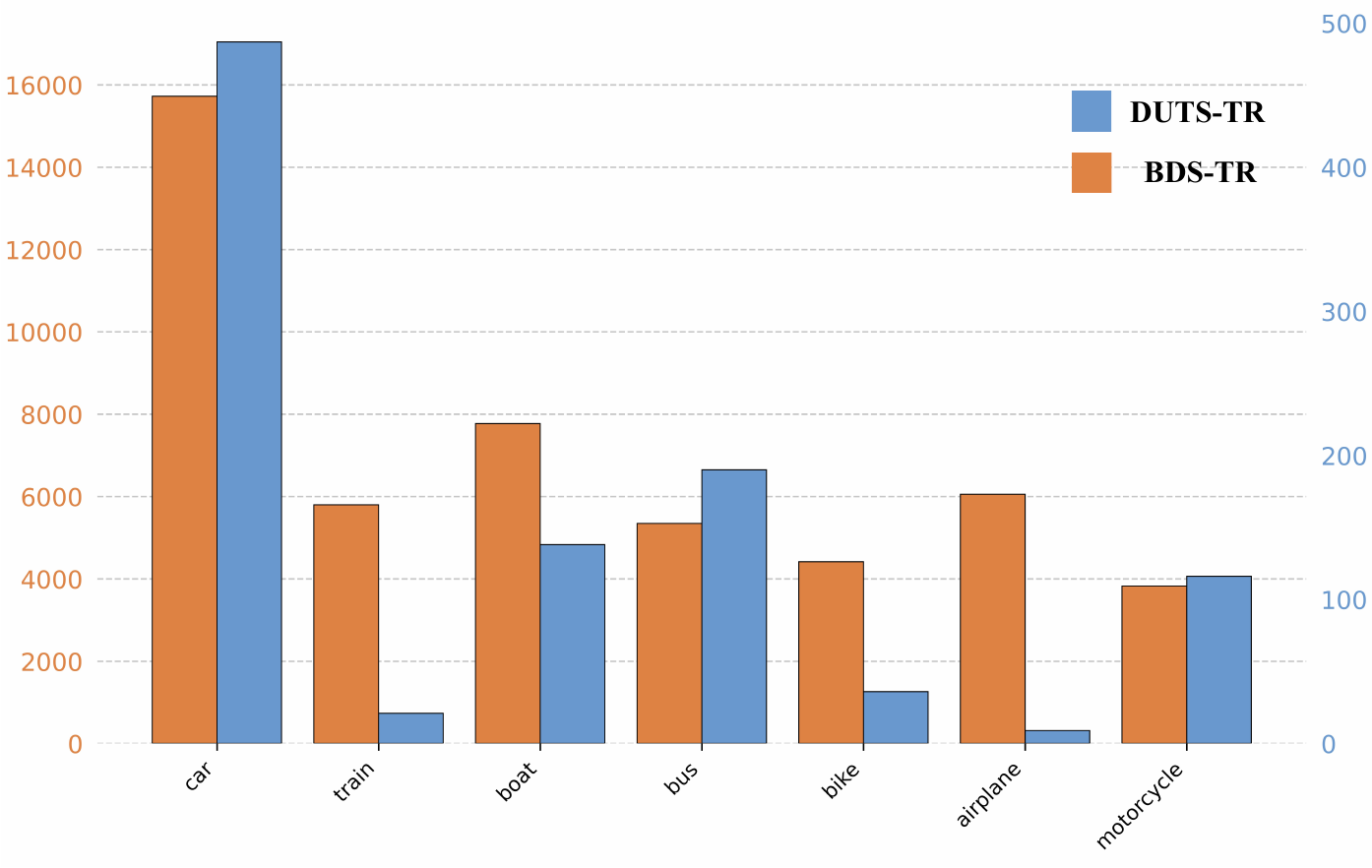}%
\label{fig:sheet3}}
\caption{Categories distribution in BDS-TR and comparison with DUTS-TR. In sub-figure (a), each point represents a category within the dataset. BDS-TR significantly surpasses DUTS-TR in both category diversity and quantity. In sub-figure (b), DUTS-TR exhibits a highly uneven distribution within the subcategory of Transportation.}
\label{fig:line_sheet}
\end{figure*}

We observed that the scenes and objects included in the original DUTS-TR dataset were relatively limited, which constrained the generalization capabilities of existing SOD algorithms in broader application scenarios. This observation motivated us to expand the DUTS-TR training set and introduce a new BDS-TR dataset. We selected suitable images from datasets such as COCO~\cite{lin2014microsoft}, OpenImages~\cite{kuznetsova2020open}, and VOC2012~\cite{everingham2015pascal}, expanding the original 10,554 images in DUTS-TR to approximately 260,000 images. Compared to DUTS-TR, our BDS-TR dataset includes a more diverse range of objects and scenes, as shown in~\cref{fig:line_chart}. Additionally, the DUTS-TR dataset exhibits an imbalance in object category distribution. In contrast, our BDS-TR dataset achieves better balance across categories, as illustrated in \cref{fig:sheet3}. This enhancement significantly improves our model’s applicability to a wider range of SOD scenarios and provides a broader foundational dataset for future SOD research.

Moreover, we propose an edge-preserving decoder based on dynamic upsampling, named DEDecoder, which effectively utilizes the edges from precise pseudo-labels to enhance performance. Inspired by Liu et al.~\cite{liu2023learning}, we progressively employ dynamic upsampling to restore feature resolution during the decoding phase. While increasing resolution, we developed a multi-scale edge-preserving module that better recovers structural information and reinforces boundary details.

The main contributions are summarized as follows:
\begin{itemize}
\item This paper introduces a novel weakly supervised SOD framework that leverages the knowledge distilled from large foundation models to generate high-precision pseudo-labels with minimal manual effort. It provides a fresh perspective for future research in SOD.

\item We introduce a new dataset for SOD called BDS-TR, which includes over 960 categories, more than 3,000 subcategories, and 260,000 images, making it substantially larger than existing benchmark datasets. This large-scale dataset significantly improves the generalization ability of SOD models, enabling them to perform well across a wider variety of scenarios.

\item We propose an edge-preserving decoder based on dynamic upsampling, which progressively restores feature resolution while preserving edge structures.

\item Evaluated on five benchmark datasets, our model sets a new benchmark by outperforming existing weakly-supervised methods and achieving state-of-the-art results comparable or superior to some of the latest fully-supervised models.

\end{itemize}

\section{Related Works}
\subsection{Salient Object Detection}
Early SOD methods primarily relied on handcrafted features and heuristic priors to segment salient objects, such as color contrast~\cite{achanta2009frequency}, background priors~\cite{yang2013saliency}, and center priors~\cite{jiang2013submodular}. Recently, with the advancements of deep learning in the field of computer vision, various architectures based on Convolutional Neural Networks (CNNs) have been employed to enhance the performance of these networks~\cite{luo2017non,zhang2017amulet,liu2018picanet}. MLMSNet~\cite{wu2019mutual} incorporates foreground boundary detection and edge supervision. AFNet~\cite{feng2019attentive} utilizes an attention feedback module to discern target structures. GateNet~\cite{zhao2020suppress} employs a gating mechanism to establish connections between features at different hierarchical levels, thereby enhancing the network's discriminative ability. The U2-Net~\cite{qin2020u2} employs a two-layer nested U-structure to capture information at various scales, thereby enabling the extraction of internal multi-resolution features without increasing computational costs. EDN~\cite{wu2022edn} effectively learns global views through extreme down-sampling while restoring object details through a scale-dependent pyramid convolution. However, the deviation from the goal of SOD lies in the fact that current benchmark datasets require pixel-level annotations, which are time-consuming and labor-intensive. This drawback has led to most existing methods being limited by the datasets. Therefore, a method for generating accurate annotations at a low cost is needed. As a result, we propose a new approach that can generate sufficient annotations with only a small amount of textual labeling required.

\begin{figure*}[!ht]
    \centering
    \includegraphics[width=\textwidth]{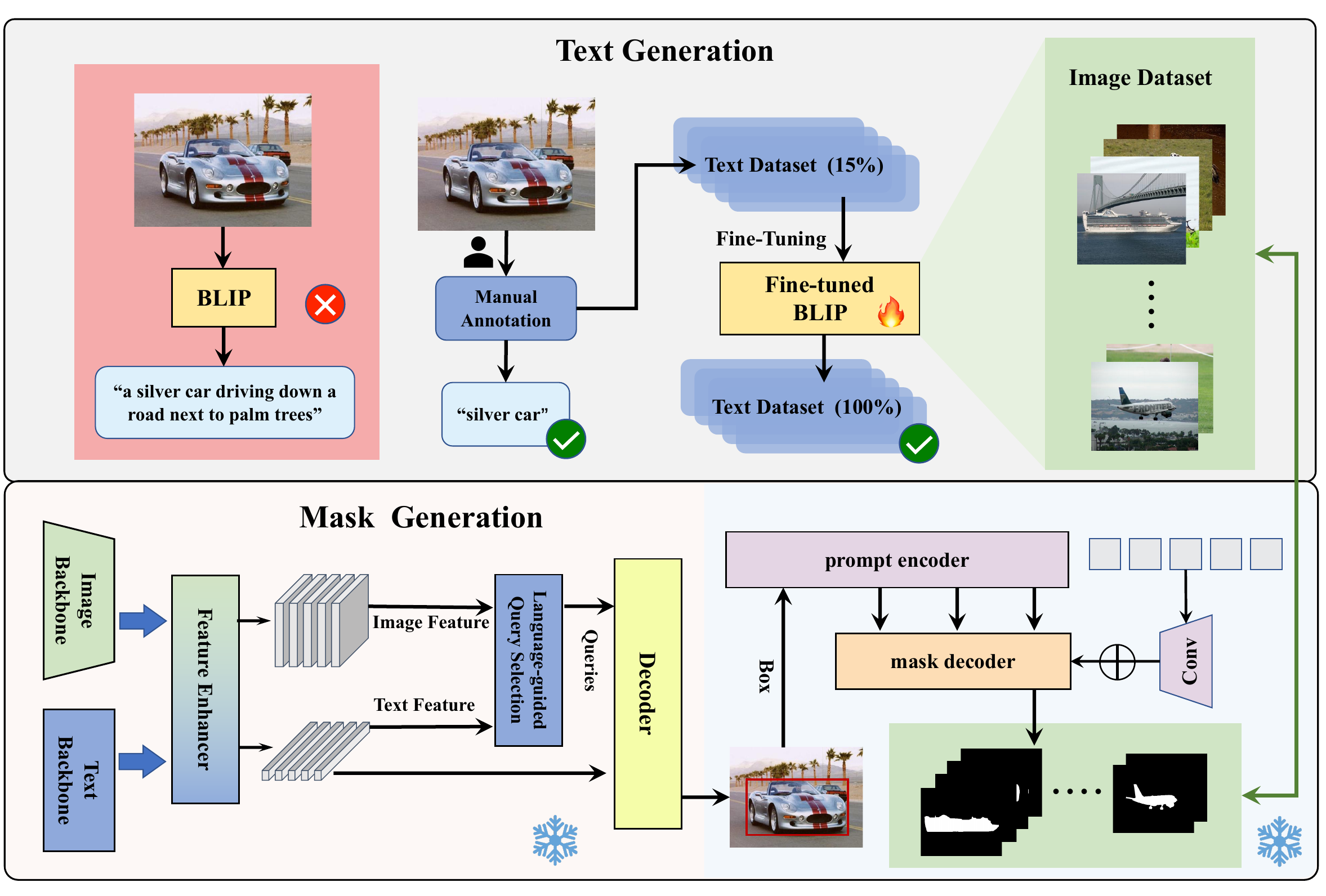}
    \caption{Annotation Pipeline, including four steps. Step 1: Manually pre-annotating a small portion of images. Step 2: Fine-tuning BLIP to generate textual descriptions for all images. Step 3: Using GroundingDINO to produce object detection boxes. Step 4: Finally employing SAM to segment the masks. The red background indicates GroundingDINO, while the blue background denotes SAM.
}
    \label{fig:anno}
\end{figure*}

\subsection{Weakly Supervised Learning}
To avoid the labor-intensive task of pixel-level annotation~\cite{feng2019attentive,wu2019mutual,wu2022edn}, weakly supervised SOD has gained increasing attention in recent years~\cite{zhang2020weakly,yu2021structure,wang2017learning,lee2021railroad}. Early weakly supervised methods focused on relatively simple approaches, such as image classification labels, bounding boxes, and scribbles. In 2017, Wang et al.~\cite{wang2017learning} introduced the first weakly supervised SOD model using image-level labels. They first trained a foreground inference network based on Fully Convolutional Networks, and then optimized the results of the first stage using an iterative Conditional Random Field (CRF) to generate pseudo-labels. Subsequent works by Li, Zheng~\cite{zheng2021weakly,li2018weakly} also proposed methods using image tags for supervision. However, these approaches struggled to distinguish multiple classes of salient objects effectively and often introduced noise from non-salient objects of the same class. Additionally, these methods require retraining the network to extract CAM maps. More importantly, image tag datasets are typically closed-set datasets, whereas SOD operates in an open-set context and is category-agnostic. This makes the use of image tags significantly limited for SOD. In 2020, Zhang~\cite{zhang2020weakly} et al. applied scribble annotations in a gated structure-aware loss and auxiliary boundary decision framework. Gao et al.~\cite{gao2022weakly} utilized point-based annotations for segmentation, introducing background suppression for non-salient regions during the second stage of training, which achieved promising results. These sparse annotations can only provide coarse annotations, lacking precise information about the details and boundaries of salient objects. Consequently, weakly supervised methods often suffer from over-segmentation or under-segmentation issues, with blurred further information leading to inaccurate detection. Therefore, improving the quality of pseudo-label details and boundaries will significantly enhance the final accuracy of the model.

\section{Methods}
\subsection{Pseudo-mask generation}
The quality of pseudo-labels is crucial for the final detection accuracy. We designed a comprehensive workflow from text to mask to generate high-quality pseudo-labels. As illustrated in \cref{fig:anno}, this process includes the following steps: manually pre-annotating a small portion of images, fine-tuning BLIP to generate textual descriptions for all images, using GroundingDINO to produce object detection boxes, and finally employing SAM to segment the masks. 

\begin{figure}[!ht]
    \centering
    \includegraphics[width=\columnwidth]{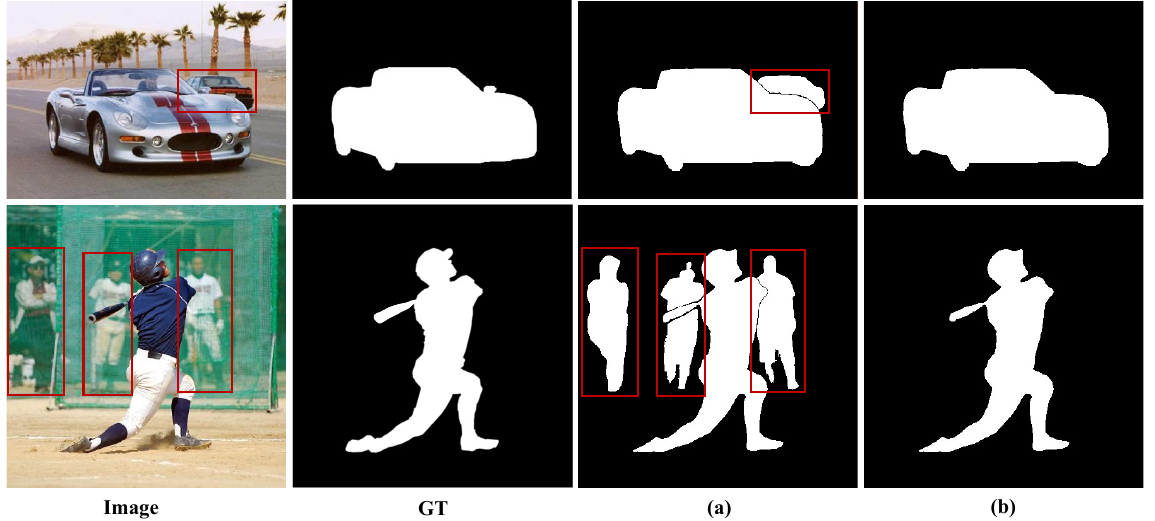}
    \caption{ Impact of Adjectives in Pseudo Masks: (a) Pseudo Labels Generated Without Adjectives (b) Pseudo Labels Generated With Adjectives. }
    \label{fig:adj}
\end{figure}

Since BLIP cannot directly focus on the salient parts of an image and instead generates captions for the entire image, it becomes challenging to create accurate pseudo-labels based on the resulting text. To address this limitation, we fine-tuned BLIP to produce more appropriate descriptions. An ideal text label should focus solely on the salient objects, which presents several challenges. First, it is necessary to exclude background elements and other distractions. Second, as SOD is not category-specific, salient objects can belong to different categories, requiring us to distinguish between salient and non-salient objects within the same category. Third, a single image may contain multiple salient objects belonging to other categories, further complicating the task. Traditional image labels have proven ineffective in addressing these issues. To overcome these challenges, we designed a structured phrase format of "(adjective) + noun", where adjectives are optional and used only when necessary to differentiate objects. We generate multiple phrases for images containing multiple salient objects, each corresponding to a specific object. This structured text representation significantly improves the accuracy of the generated pseudo-labels, as illustrated in \cref{fig:adj}. Furthermore, this adjective-based specification maintains the simplicity of the text, facilitating subsequent BLIP fine-tuning and object detection using GroundingDINO without introducing unnecessary complexity.

We manually annotated approximately 15\%  of the images in the DUTS-TR dataset, which contains 10,554 images. We then fine-tuned Blip using this manually annotated data. After fine-tuning, we used the model to generate textual descriptions for the entire dataset. When generating a textual label, we employ GroundingDINO to detect objects within images. For each detected object, a bounding box $b_i$ is obtained, represented by a quadruple $(x_1,y_1,x_2,y_2)$, where $x_1,y_1$ denote the coordinates of the top-left corner, and $x_2,y_2$ signify the bottom-right corner coordinates of the box. Thus, for an image $\mathcal{I}$ and text $\mathcal{T}$, we generate a set of bounding boxes $\displaystyle \mathcal{B}_\mathcal{I}=b_1^{logit_1} \cup b_2^{logit_2} \cup \cdots \cup b_n^{logit_n}$, the term $logit_i$ denotes the confidence score associated with bounding box $b_i$. Upon generating the bounding boxes, we feed these boxes into SAM to acquire object segmentation masks. Some examples of the high-quality pseudo-labels generated by this process are shown in ~\cref{fig:pseudo}.

\begin{figure}[!ht]
    \centering
    \includegraphics[width=\columnwidth]{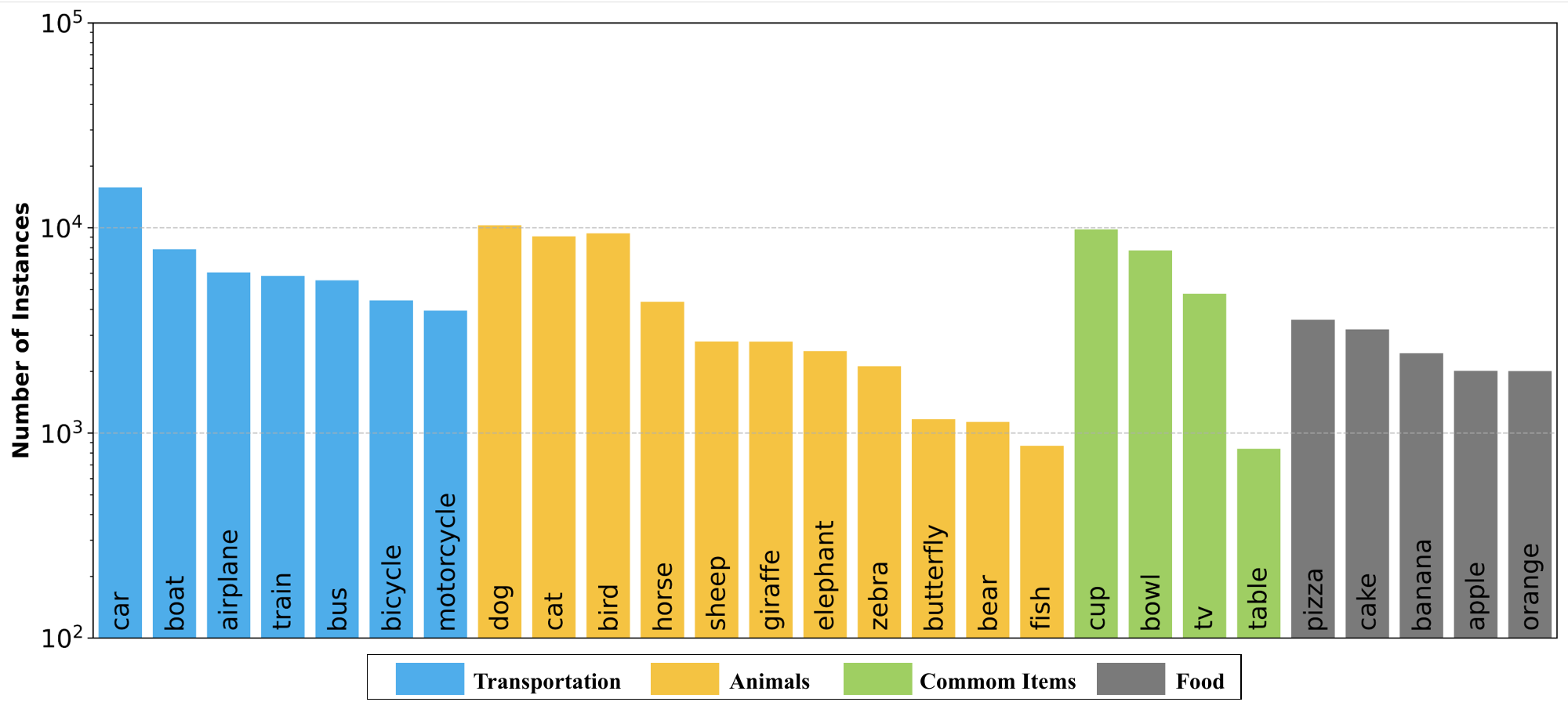}
    \caption{Histogram of Some Common Parent and Subcategories, with Objects Sorted by Frequency. The entire dataset consists of various objects typically found in everyday scenarios.}
    \label{fig:sheet}
\end{figure}

\subsection{BDS-TR}
Considering the wide range of application scenarios for SOD and the diversity of object categories involved, we expanded the benchmark dataset DUTS-TR to accommodate more application scenarios and provide more data for future research. As mentioned in the previous section, generating pseudo-labels requires minimal manual effort, making it feasible to expand a large-scale dataset. We first selected appropriate images for salient object detection from COCO, OpenImages, and VOC2012 datasets. Since these datasets are not specifically designed for SOD tasks, we randomly sampled images and manually filtered them to construct a dataset tailored to salient object detection scenarios. Compared to the previous DUTS-TR dataset, which contained only around 300 object categories, the expanded dataset has significantly improved the diversity and quantity of object types. Specifically, the expanded dataset includes approximately 960 major object categories, covering a wide range of items from everyday objects to those in complex scenes, as well as over 3,000 subcategories, further refining the features of these objects. The distribution details of some parent categories and their subclasses are shown in ~\cref{fig:sheet}. This diversity in both major and minor categories allows our dataset to cover a broader array of real-world SOD scenarios. Moreover, the expanded dataset not only achieves diversity in categories but also significantly increases in size, with more than 260,000 images, far exceeding the scale of the original DUTS-TR dataset.

After creating the image set, we manually annotated approximately 8,000 images with text labels, accounting for about 3\% of the total. We then generated pseudo-labels using the process above, followed by a final round of manual inspection to further improve the quality of our dataset.

\begin{figure*}[!ht]
    \centering
    \includegraphics[width=\textwidth]{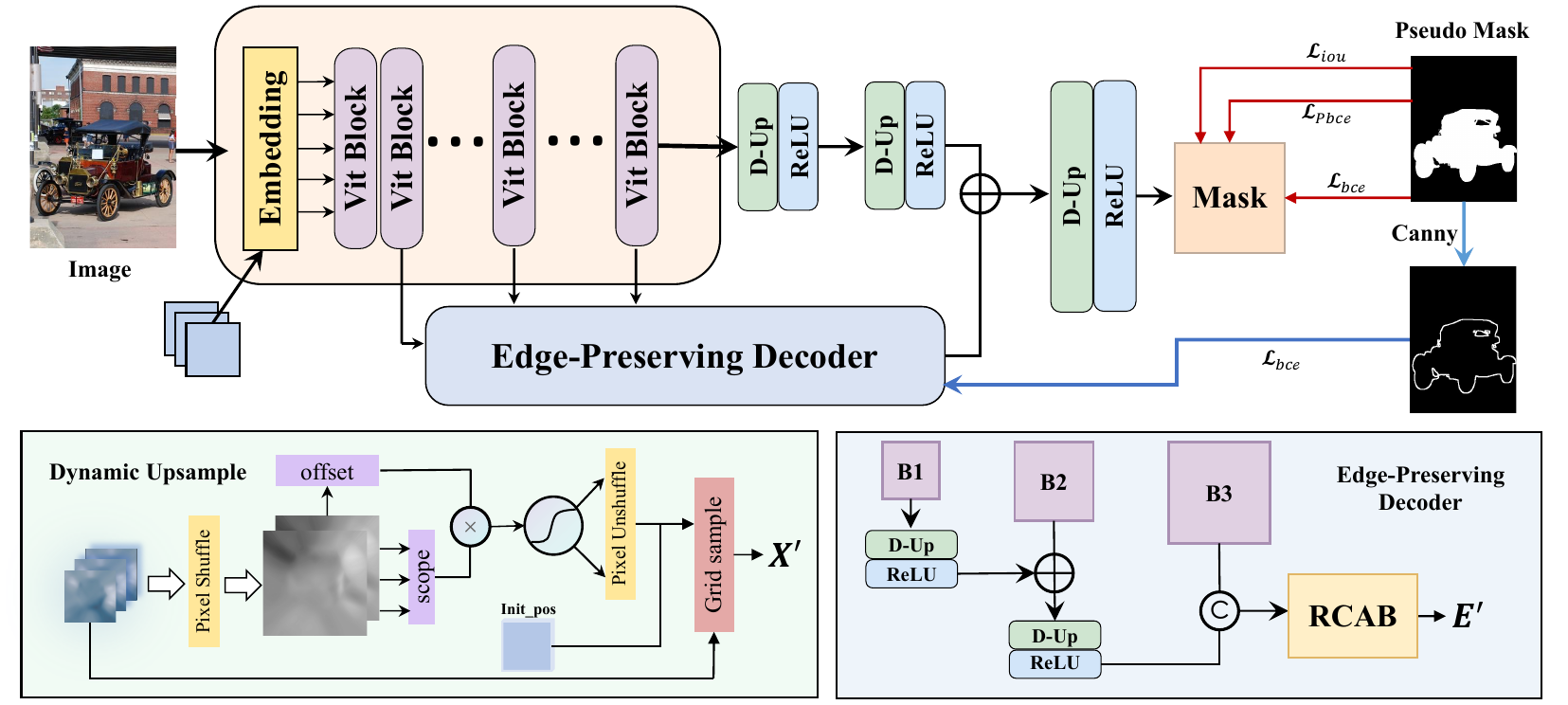}
    \caption{Structure of the Dynamic Edge-Preserving Decoder.}
    \label{fig:decoder}
\end{figure*}

\subsection{Network and edge decoder}

Considering the exceptional performance of Vision Transformers (ViT)~\cite{dosovitskiy2020image} in various visual tasks and their superior global information processing capability compared to ResNet~\cite{he2016deep}, we select DINO ViT as our backbone network. DINO transformer~\cite{caron2021emerging} leverages self-supervised learning to train vision transformers (ViTs) without additional annotated information, exhibiting unique capabilities in the explicit representation of image semantic segmentation. It possesses enhanced semantic expression and discrimination abilities, making it more suitable for weakly supervised tasks~\cite{kang2023distilling,gomel2023box}. Specifically, a series of image tokens are generated for an input image of size \((3, H, W)\). These patches are linearly embedded into \(P \in  \mathbb{R}^{C \times  \frac{W}{8} \times  \frac{H}{8}}\), where \(C\) denotes the token feature dimensionality.

Inspired by Liu et al.~\cite{liu2023learning}, we implemented dynamic upsampling within our decoder to restore image resolution effectively. This approach interpolates input features to create a continuous feature map using bilinear interpolation. Subsequently, we generate content-aware sampling points to resample this continuous feature map. The point-wise offsets are obtained through a linear projection, and the corresponding values at these offset locations are computed through a resampling process based on grid interpolation. This method does not require high-resolution guiding features as input, simplifying the process and reducing computational costs.

Edge detection significantly enhances WSOD performance by restoring structural details~\cite{gao2022weakly,zhang2020weakly}. Our edge-preserving decoder receives features extracted from the ViT blocks, denoted as \( f = \{f_i \mid i = 1, 2, 3\} \). The first feature \( f_1 \) undergoes dynamic upsampling, ReLU activation~\cite{glorot2011deep}, and Batch Normalization (BN)~\cite{ioffe2015batch}, then is added to the second feature \( f_2 \). After another dynamic upsampling, the result is concatenated with the third feature \( f_3 \). This concatenated output \( f_e \) is processed by a Residual Channel Attention Block (RCAB)~\cite{zhang2018image} to suppress non-edge information. We utilize the Canny~\cite{canny1986computational} edge detector to perform edge detection on the pseudo labels. The edge supervision is achieved using binary cross-entropy (BCE) loss. Due to the high quality of the pseudo labels, accurate edge information can be generated, significantly enhancing the edge-preserving decoder's performance.

\subsection{Loss Function}
In our network, we employ binary cross-entropy, partial cross-entropy loss~\cite{tang2018normalized}, and Intersection over Union (IoU) loss~\cite{mattyus2017deeproadmapper}:

\begin{multline}
\mathcal{L}_{bce}= -\sum_{i=1}^H \sum_{j=1}^W G(i, j) \log (\operatorname{Y}(i, j)) \\
+(1-G(i, j)) \log (1-\operatorname{Y}(i, j)),
\end{multline} where \(H\) and \(W\) denote the height and width of the images, respectively. \(G(i, j)\) represents the ground truth label, while \(Y(i, j)\) denotes the predicted label for the pixel at location \((i, j)\). The Partial Binary Cross-Entropy Loss is employed to focus solely on certain regions while disregarding areas of uncertainty:

\begin{equation}
\mathcal{L}_{{pbce }}=- \frac{|N|}{|J|} \sum_{j \in J} \left[g_j \log \left(y_j\right)+\left(1-g_j\right) \log \left(1-y_j\right)\right],
\end{equation} In our loss function, to better capture the target structure, we also incorporate the Intersection over Union (IoU) loss:

\begin{equation}
\mathcal{L}_{{IoU}} = \frac{\sum_{i=1}^H \sum_{j=1}^W \mathcal{I}(G(i, j) \cap Y(i, j))}{\sum_{i=1}^H \sum_{j=1}^W \mathcal{U}(G(i, j) \cup Y(i, j))},
\end{equation} Here, \(\mathcal{I}\) denotes the area of the intersection region, that is, the count of pixels for which both \(G(i, j)\) and \(Y(i, j)\) are true; \(\mathcal{U}\) represents the area of the union region, that is, the count of pixels where at least one of \(G(i, j)\) or \(Y(i, j)\) is true. The overall loss of our training can be formulated as:

\begin{equation}
\mathcal{L}_{final} = \alpha_1\mathcal{L}_{bce} + \alpha_2\mathcal{L}_{pbce} + \alpha_3\mathcal{L}_{IoU},
\end{equation} where $\alpha_1,\alpha_2,\alpha_3$ are the weights. In our experiments, they are all set to 1.

\begin{table*}[!ht]
  \centering
\renewcommand\tabcolsep{3.5pt}
  \caption{Experiments for SOD benchmark datasets measured in mean F-measure$\left(\mathcal{F}_\beta\right)$,E-measure$\left(\mathcal{E}_\xi\right)$, S-measure$\left(\mathcal{S}\right)$, max F-measure$\left(\mathcal{F}_{m}\right)$ and MAE$\left(\mathcal{M}\right)$. $\uparrow$ and $\downarrow$ indicate that the larger and smaller scores are better, respectively. The top two results are highlight in {\textcolor{red}{\textbf{red}}}, and {\textcolor{blue}{\textbf{blue}}}.}
  \label{tab:benchmark}
  \renewcommand\tabcolsep{2.0pt}
  \renewcommand{\arraystretch}{1.7}
  \resizebox{\textwidth}{!}{
   \scriptsize
    \begin{tabularx}{\textwidth}{l|c|XXXXX|XXXXX|XXXXX|XXXXX|XXXXX}
    \toprule
          \multirow{2}[1]{*}{Methods} & \multirow{2}[1]{*}{Year} & \multicolumn{5}{c|}{ECSSD} & \multicolumn{5}{c|}{DUT-O} & \multicolumn{5}{c|}{PASCAL-S} & \multicolumn{5}{c|}{DUTS-TE} & \multicolumn{5}{c}{HKU-IS} \\
          &      & $\mathcal{S} \uparrow$ & $\mathcal{F}_{m} \mkern-7mu \uparrow$ & $\mathcal{F}_\beta \uparrow$ & $\mathcal{E}_{\xi} \uparrow$ & $\mathcal{M} \downarrow$     & $\mathcal{S} \uparrow$ & $\mathcal{F}_{m} \mkern-7mu \uparrow$ & $\mathcal{F}_\beta \uparrow$ & $\mathcal{E}_{\xi} \uparrow$ & $\mathcal{M} \downarrow$ & $\mathcal{S} \uparrow$ & $\mathcal{F}_{m} \mkern-7mu \uparrow$ & $\mathcal{F}_\beta \uparrow$ & $\mathcal{E}_{\xi} \uparrow$ & $\mathcal{M} \downarrow$ & $\mathcal{S} \uparrow$ & $\mathcal{F}_{m} \mkern-7mu \uparrow$ & $\mathcal{F}_\beta \uparrow$ & $\mathcal{E}_{\xi} \uparrow$ & $\mathcal{M} \downarrow$ & $\mathcal{S} \uparrow$ & $\mathcal{F}_{m} \mkern-7mu\uparrow$ & $\mathcal{F}_\beta \uparrow$ & $\mathcal{E}_{\xi} \uparrow$ & $\mathcal{M} \downarrow$\\
    \midrule
    \multicolumn{27}{c}{Fully Sup.Methods} \\
    \midrule   
    BASNet \cite{qin2019basnet}     & 2019 & .916 & .942 & .930 & .921 & .037 & .836 & .805 & .793 & .869 & .057 & .838 & .853 & .837 & .852 & .076 & .865 & .859 & .845 & .884 & .048 & .908 & .928 & .914 & .945 & .032 \\
    F3Net\cite{wei2020f3net}        & 2020 & .924 & .945 & .933 & .946 & .033 & .838 & .813 & .794 & .876 & .053 & .861 & .871 & .852 & .894 & .062 & .888 & .891 & .867 & .918 & .036 & .917 & .936 & .918 & .958 & .028\\
    LDF\cite{wei2020label}          & 2020 & .924 & .950 & .937 & .950 & .034 & .839 & .819 & .801 & .881 & .052 & .862 & .874 & .857 & {\textcolor{blue}{\textbf{.904}}} & .060 & .892 & .897 & {\textcolor{blue}{\textbf{.878}}} & {\textcolor{blue}{\textbf{.929}}} & .034 & .919 & .939 & .922 & .959 & .028 \\
    MINet\cite{pang2020multi}       & 2020 & .924 & .947 & .930 & .953 & .034 & .832 & .809 & .789 & .873 & .056 & .856 & .866 & .846 & .898 & .063 & .884 & .883 & .860 & .917 & .037 & .918 & .934 & .916 & .960 & .028 \\
    Gate\cite{zhao2020suppress}     & 2020 & .919 & .945 & .925 & .943 & .040 & .838 & .818 & .791 & .868 & .055 & .858 & .869 & .845 & .884 & .067 & .885 & .887 & .855 & .902 & .040 & .915 & .933 & .909 & .953 & .033 \\
    EDN\cite{wu2022edn}             & 2022 & .926 & .951 & .939 & {\textcolor{red}{\textbf{.955}}} & .032 & {\textcolor{blue}{\textbf{.849}}} & .828 & .813 & .885 & .049 & .864 & .879 & .865 & {\textcolor{blue}{\textbf{.904}}} & .061 & .892 & .894 & {\textcolor{blue}{\textbf{.878}}} & .928 & .035 & .924 & .941 & .927 & {\textcolor{blue}{\textbf{.962}}} & .026 \\
    MENet\cite{wang2023pixels}      & 2023 & {\textcolor{blue}{\textbf{.927}}} & {\textcolor{blue}{\textbf{.954}}} & {\textcolor{blue}{\textbf{.941}}} & {\textcolor{blue}{\textbf{.954}}} & {\textcolor{blue}{\textbf{.030}}} & {\textcolor{blue}{\textbf{.849}}} & {\textcolor{blue}{\textbf{.833}}} & {\textcolor{blue}{\textbf{.817}}} & {\textcolor{blue}{\textbf{.891}}} & {\textcolor{blue}{\textbf{.045}}} & {\textcolor{blue}{\textbf{.872}}} & {\textcolor{blue}{\textbf{.889}}} & {\textcolor{blue}{\textbf{.869}}} & {\textcolor{red}{\textbf{.913}}} & {\textcolor{blue}{\textbf{.054}}} & {\textcolor{blue}{\textbf{.904}}} & {\textcolor{blue}{\textbf{.913}}} & {\textcolor{red}{\textbf{.893}}} & {\textcolor{red}{\textbf{.942}}} &  {\textcolor{blue}{\textbf{.028}}} & {\textcolor{blue}{\textbf{.927}}} & {\textcolor{red}{\textbf{.948}}} & {\textcolor{red}{\textbf{.931}}} & {\textcolor{red}{\textbf{.965}}} & {\textcolor{red}{\textbf{.023}}} \\
    SelfRe\cite{yun2023towards}     & 2024 & {\textcolor{red}{\textbf{.935}}} & {\textcolor{red}{\textbf{.957}}} & {\textcolor{red}{\textbf{.943}}} & .936 & {\textcolor{red}{\textbf{.027}}} & {\textcolor{red}{\textbf{.860}}} & {\textcolor{red}{\textbf{.836}}} & {\textcolor{red}{\textbf{.819}}} & {\textcolor{red}{\textbf{.892}}} & {\textcolor{red}{\textbf{.043}}} & {\textcolor{red}{\textbf{.880}}} & {\textcolor{red}{\textbf{.894}}} & {\textcolor{red}{\textbf{.875}}} & .882 & {\textcolor{red}{\textbf{.051}}} & {\textcolor{red}{\textbf{.910}}} & {\textcolor{red}{\textbf{.916}}} & {\textcolor{red}{\textbf{.893}}} & .924 & {\textcolor{red}{\textbf{.027}}} & {\textcolor{red}{\textbf{.930}}} & {\textcolor{blue}{\textbf{.947}}} & {\textcolor{blue}{\textbf{.928}}} & .960 & {\textcolor{blue}{\textbf{.024}}} \\

    \midrule
    \multicolumn{27}{c}{Weakly Sup./Unsup. Methods} \\
    \midrule
    
    WSSA\cite{zhang2020weakly}      & 2020 & .865 & .888 & .880 & .917 & .059 & .784 & .753 & .737 & .844 & .068 & .797 & .808 & .795 & .856 & .092 & .803 & .788 & .772 & .868 & .062 & .864 & .880 & .870 & .932 & .047 \\
    MFNet\cite{piao2021mfnet}       & 2021 & .822 & .861 & .843 & .877 & .092 & .713 & .651 & .628 & .756 & .113 & .767 & .782 & .764 & .810 & .119 & .768 & .752 & .727 & .815 & .088 & .834 & .862 & .839 & .903 & .067 \\
    SCW\cite{yu2021structure}       & 2021 & .881 & .914 & .909 & .931 & .049 & .811 & .782 & .777 & {\textcolor{blue}{\textbf{.869}}} & {\textcolor{blue}{\textbf{.060}}} & .819 & .840 & .835 & .880 & .078 & .840 & .844 & .839 & {\textcolor{blue}{\textbf{.906}}} & .049 & .882 & .908 & .903 & .942 & .038 \\
    A2S\cite{zhou2022activation}    & 2022 & .866 & .904 & .900 & .910 & .064 & .795 & .775 & .753 & .840 & .068 & .787 & .811 & .806 & .837 & .103 & .810 & .806 & .781 & .860 & .064 & .882 & .905 & .896 & .937 & .042 \\
    TSD\cite{zhou2023texture}       & 2023 & .893 & .926 & .922 & {\textcolor{blue}{\textbf{.940}}} & .044 & .812 & .788 & .773 & .863 & {\textcolor{blue}{\textbf{.060}}} & .830 & .846 & .836 & .886 & .072 & .842 & .842 & .831 & .902 & .047 & .889 & .914 & .909 & .947 & .037\\
    PSOD\cite{gao2022weakly}        & 2022 & {\textcolor{blue}{\textbf{.912}}} & {\textcolor{blue}{\textbf{.936}}} & {\textcolor{blue}{\textbf{.927}}} & {\textcolor{red}{\textbf{.953}}} & {\textcolor{blue}{\textbf{.038}}} & {\textcolor{blue}{\textbf{.829}}} & {\textcolor{blue}{\textbf{.813}}} & {\textcolor{blue}{\textbf{.791}}} & .864 & .062 & {\textcolor{blue}{\textbf{.852}}} & {\textcolor{blue}{\textbf{.864}}} & {\textcolor{blue}{\textbf{.851}}} & {\textcolor{red}{\textbf{.894}}} & {\textcolor{blue}{\textbf{.066}}} & {\textcolor{blue}{\textbf{.853}}} & {\textcolor{blue}{\textbf{.857}}} & {\textcolor{blue}{\textbf{.842}}} & .903 & {\textcolor{blue}{\textbf{.046}}} & {\textcolor{blue}{\textbf{.900}}} & {\textcolor{blue}{\textbf{.925}}} & {\textcolor{blue}{\textbf{.915}}} & {\textcolor{blue}{\textbf{.955}}} & {\textcolor{blue}{\textbf{.034}}} \\
    \bottomrule
    Ours                            &  -   & {\textcolor{red}{\textbf{.933}}} & {\textcolor{red}{\textbf{.958}}} & {\textcolor{red}{\textbf{.951}}} & .937 & {\textcolor{red}{\textbf{.028}}} & {\textcolor{red}{\textbf{.860}}} & {\textcolor{red}{\textbf{.841}}} & {\textcolor{red}{\textbf{.830}}} & .{\textcolor{red}{\textbf{899 }}}& {\textcolor{red}{\textbf{.045}}} & {\textcolor{red}{\textbf{.883}}} & {\textcolor{red}{\textbf{.894}}} & {\textcolor{red}{\textbf{.881}}} & {\textcolor{blue}{\textbf{.889}}} & .{\textcolor{red}{\textbf{051 }}}& {\textcolor{red}{\textbf{.917}}} & {\textcolor{red}{\textbf{.923}}} & {\textcolor{red}{\textbf{.912}}} & {\textcolor{red}{\textbf{.937}}} & {\textcolor{red}{\textbf{.026}}} & {\textcolor{red}{\textbf{.933}}} & {\textcolor{red}{\textbf{.950}}} & {\textcolor{red}{\textbf{.941}}} & {\textcolor{red}{\textbf{.965}}} & {\textcolor{red}{\textbf{.023}}} \\
    \bottomrule
    \end{tabularx}%
    }
    \vspace{-0.05in}
\end{table*}%

\begin{figure*}[ht]
    \centering
    \includegraphics[width=\textwidth]{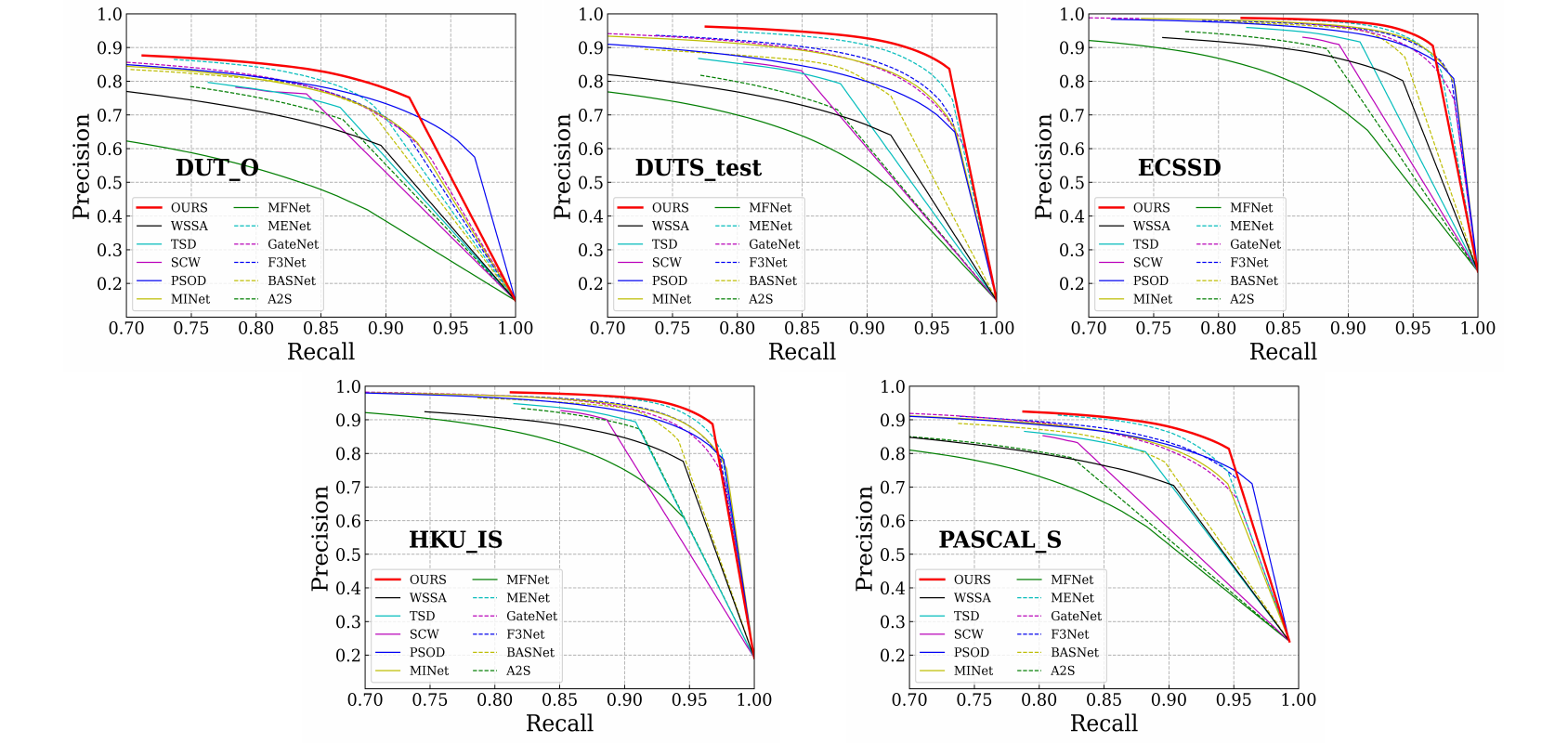}
    \caption{ Illustration of precision-recall curves.}
    \label{fig:PRcurve}
\end{figure*}

\section{Experiments}
\subsection{Implementation Details}
In the initial phase of pseudo label generation, we employ Grounding-DINO~\cite{liu2023grounding} for locating targets described by text and SAM~\cite{kirillov2023segment} for segmenting the located targets. Specifically, we utilize the Swin-B Grounding-DINO model and the ViT-H SAM model. During the second phase of model training, our proposed model is implemented on the PyTorch framework, using T-DUTS as the training dataset. Training is conducted using two NVIDIA GeForce RTX 3090 GPUs. 

We employ ViT-small~\cite{caron2021emerging} as our backbone network, with the pre-trained model being self-supervised trained on ImageNet-1k~\cite{russakovsky2015imagenet} using DINO, comprising 21 million parameters. To prevent overfitting, we apply image augmentation techniques such as random flipping.

The patch size for DINO ViT-S is set to 8, with a dropout rate of 0.1~\cite{srivastava2014dropout} and a gradient clipping value of 0.5. We use AdamW~\cite{loshchilov2017fixing} as the optimizer, with an initial learning rate of 1e-5, decaying through a poly schedule~\cite{zhao2017pyramid} and employing a warm-up strategy during the first 12,000 iterations. The input images are resized to 352$\times$352, with a batch size of 8 and 60 epochs. For testing, each image is resized to the respective resolution and then fed into our network to obtain saliency predictions without any post-processing.

\begin{figure*}[!ht]
    \centering
    \includegraphics[width=\textwidth]{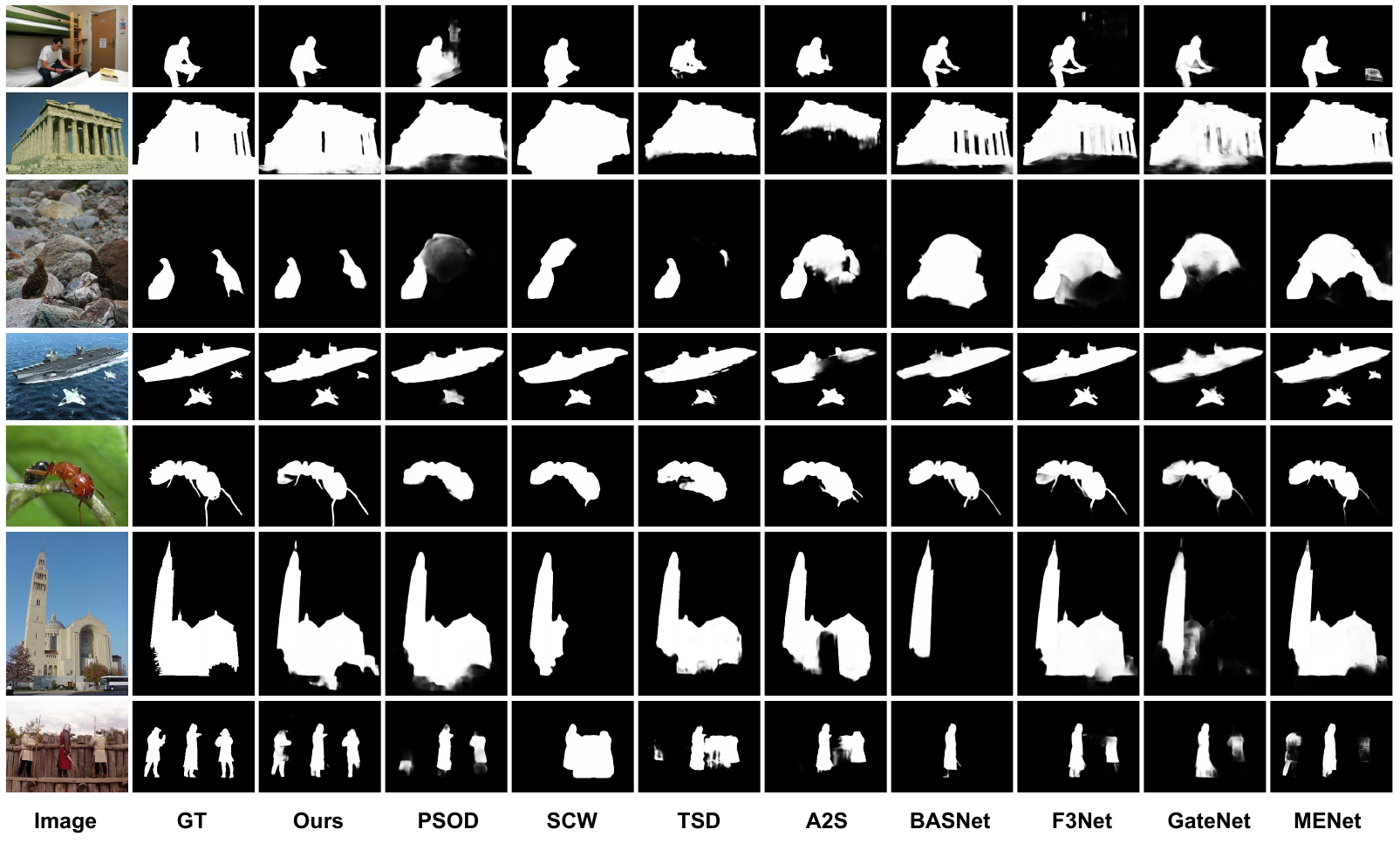}
    \caption{Qualitative comparison with different methods. Saliency maps produced by our model are clearer and more accurate than those of other methods in various challenging scenarios.
}
    \label{fig:result}
\end{figure*}

\subsection{Evaluation Datasets and Metrics}
\subsubsection{Datasets.}
We evaluate the performance of our model on five publicly available benchmark datasets: ECSSD~\cite{yan2013hierarchical}, PASCAL-S~\cite{li2014secrets}, DUT-O~\cite{yang2013saliency}, HKU-IS~\cite{li2015visual}, and DUTS-test. The DUTS~\cite{wang2017learning} dataset comprises 10,553 training images (DUTS-TR) and 5,019 test images (DUTS-TE). ECSSD features 1,000 images characterized by structurally complex natural scenes. HKU-IS includes 4,447 images of complex scenarios with multiple salient objects. DUT-O encompasses 5,168 images with intricate backgrounds, while  PASCAL-S comprises 850 challenging photos.

\subsubsection{Evaluation Metrics.}
To comprehensively and fairly evaluate various methods, we employed six metrics: precision-recall curve, mean absolute error $\left(\mathcal{M}\right)$~\cite{borji2015salient}, E-measure$\left(\mathcal{E}_\xi\right)$~\cite{fan2018enhanced}, S-measure$\left(\mathcal{S}\right)$~\cite{fan2017structure}, mean F-measure$\left(\mathcal{F}_\beta\right)$~\cite{achanta2009frequency} and max F-measure$\left(\mathcal{F}^{max}\right)$~\cite{achanta2009frequency}. Specifically, the F-measure assesses overall performance by considering region similarity. The Mean Absolute Error quantifies the average absolute disparity between the saliency map and ground truth. The E-measure uniquely leverages image and local pixel-level statistics for evaluating binary saliency maps. The S-measure is an indicator extensively used to measure the structural similarity between the original and the detected images. The Precision-Recall curve is essential for evaluating the precision of optimistic predictions alongside the model's ability to detect all relevant entities in the dataset.

\subsection{Compare with State-of-the-arts}
\subsubsection{Quantitative Comparison}

Our approach is benchmarked against state-of-the-art models, as shown in \cref{tab:benchmark} and \cref{fig:PRcurve}. To ensure a fair comparison, saliency results for the various methods were obtained directly from the authors or through their publicly available code. The top two results are highlighted in red and blue for clarity. As \cref{tab:benchmark} demonstrates, our method achieves significant performance improvements over current state-of-the-art models in weakly-supervised learning. Notably, for the MAE metric, it outperforms existing methods by 26.3\%, 25\%, 22.7\%, 43.4\%, and 32.3\% across the five datasets. For other metrics that assess image structure, such as $\mathcal{S}$ and $\mathcal{E}$, our method also demonstrates significant advantages. This can be primarily attributed to the precise structure and edges of our pseudo-labels and our proposed DEDecoder's effectiveness. \Cref{fig:PRcurve} further illustrates the superior performance of our approach through a quantitative comparison, highlighting a substantial margin of improvement over previous methods. Moreover, as presented in \cref{tab:benchmark}, our approach matches or surpasses fully supervised models' performance. This indirectly indicates that the pseudo-labels generated by our method achieve a quality comparable to ground-truth labels.

\subsubsection{Qualitative Evaluation}
To illustrate the efficacy of our methodology, we have visualized a comparative analysis between our framework and various state-of-the-art models as depicted in \cref{fig:result}. Our framework outperforms weakly-supervised approaches and sometimes even surpasses fully-supervised models in saliency map quality. For instance, as shown in rows 2 and 5, our approach accurately captures intricate structures and edges. Our method effectively avoids the missed detections common in other models in scenes containing multiple salient objects, such as rows 4 and 7. Rows 1 and 6 highlight our ability to suppress the interference of non-salient objects and occlusions, ensuring robust saliency detection. Notably, in camouflaged scenes where the foreground shares similar color and texture patterns with the background (row 3), our method successfully identifies the camouflaged regions, demonstrating its exceptional adaptability and accuracy.

\section{Ablation Study}


\subsection{Influence of Text.}
In generating pseudo-labels through text, we incorporated adjectives to specify and delineate salient objects more effectively, as illustrated in \cref{fig:adj}. The use of adjectives plays a crucial role in improving pseudo-label precision, which in turn has a direct and noticeable impact on the model's overall accuracy. To rigorously evaluate the contribution of adjective-based text, we conducted ablation experiments as detailed in \cref{tab:adj}. The first row presents the results obtained without using adjectives in the text, while the second row reflects the results with adjectives incorporated as part of the object specification process. The difference between the two sets of results underscores the importance of adjectives in enhancing label quality.

\begin{figure}[!ht]
    \centering
    \includegraphics[width=\columnwidth]{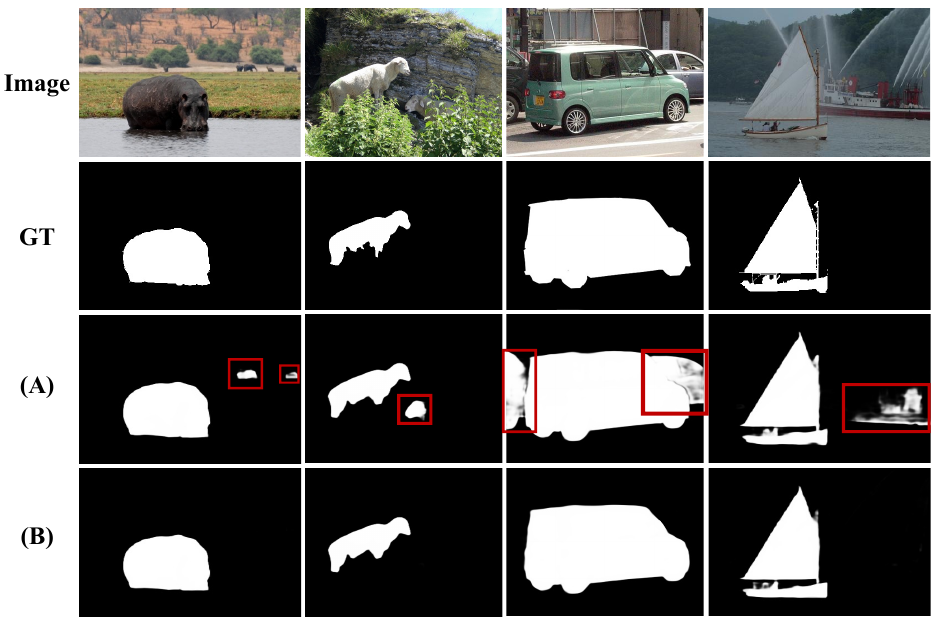}
    \caption{ Impact of Adjectives in Predict Saliency Maps. }
    \label{fig:adj_predmap}
\end{figure}

Moreover, to provide a more intuitive understanding of how different text-based specifications influence the generated pseudo-labels, we have visualized the results in \cref{fig:adj_predmap}. The comparison between Row A (without adjectives) and Row B (with adjectives) shows that pseudo-labels produced through adjective-based descriptions more effectively emphasize salient objects. In Row B, the objects of interest are highlighted with greater clarity, and much of the noisy background is removed, demonstrating a substantial reduction in unwanted interference. This visual evidence and the quantitative results illustrate the tangible benefits of integrating adjectives into the pseudo-label generation process, leading to improved model performance and more reliable object detection.

 \begin{table}[!b]
  \centering
  \caption{Comparison with different texts.}
  \label{tab:adj}
\renewcommand\tabcolsep{2.0pt}
  \renewcommand{\arraystretch}{1.33}
  \vspace{-0.1in}
  \begin{tabular}{c|cccc|cccc|cccc}
    \toprule
    \toprule
    \multirow{2}[1]{*}{} & \multicolumn{4}{c|}{DUT-O} & \multicolumn{4}{c|}{ECSSD}  & \multicolumn{4}{c}{PASCAL-S}  \\
   & $\mathcal{S} \uparrow$ & $\mathcal{F}_{m} \mkern-7mu \uparrow$ & $\mathcal{E}_{\xi} \uparrow$ & $\mathcal{M} \downarrow$ 
   & $\mathcal{S} \uparrow$ & $\mathcal{F}_{m} \mkern-7mu \uparrow$ & $\mathcal{E}_{\xi} \uparrow$ & $\mathcal{M} \downarrow$
   & $\mathcal{S} \uparrow$ & $\mathcal{F}_{m} \mkern-7mu \uparrow$ & $\mathcal{E}_{\xi} \uparrow$ & $\mathcal{M} \downarrow$
    \\
    \midrule
    w/o & .850 & .828 & .885 & .052 & .932 & .952 & .930 & .030 & .883 & .894 & .884 & .051 \\
    w/  & \textbf{.860} & \textbf{.841} & \textbf{.899} & \textbf{.045} & \textbf{.933} & \textbf{.958} & \textbf{.937} & \textbf{.028} & \textbf{.883} & \textbf{.894} & \textbf{.889} & \textbf{.050} \\
    \bottomrule
    \bottomrule
  \end{tabular}
  \vspace{-0.1in}
\end{table}

\subsection{Influence of Dataset. }
We conducted experiments using BDS-TR, and the results are summarized in ~\cref{tab:BDS}. Since we cannot provide precise annotations for the expanded test set, we continued using previous benchmark test sets for comparison. While this approach does not fully showcase the comprehensive advantages of our BDS-TR dataset in terms of object diversity and scene complexity, the results are nonetheless revealing. On the relatively simple ECSSD dataset, the performance improvements achieved by BDS-TR are modest. However, on more challenging datasets like DUT-O and PASCAL-S, BDS-TR substantially enhances model performance, far exceeding the results of fully supervised models shown in ~\cref{tab:benchmark}. This significant leap in performance highlights the power of BDS-TR in handling complex and diverse scenes.

What sets BDS-TR apart is its extensive coverage of various objects and scenes, something that previous datasets like DUTS-TR simply could not offer. By expanding the dataset, BDS-TR provides a richer and more detailed representation of real-world scenarios. This diversity translates directly into better model generalization across different contexts. As demonstrated by the remarkable improvements on more difficult datasets, BDS-TR enhances the model’s ability to handle complex tasks, outperforming fully supervised approaches. In essence, BDS-TR offers a powerful foundation for future SOD models by equipping them with a broader, more representative training dataset that drives better generalization and robustness.

 \begin{table}[!ht]
  \centering
  \caption{Comparison of Results Using DUTS-TR and BDS-TR}
  \label{tab:BDS}
\renewcommand\tabcolsep{2.0pt}
  \renewcommand{\arraystretch}{1.33}
  \vspace{-0.1in}
  \begin{tabular}{c|cccc|cccc|cccc}
    \toprule
    \toprule
    \multirow{2}[1]{*}{} & \multicolumn{4}{c|}{DUT-O} & \multicolumn{4}{c|}{ECSSD}  & \multicolumn{4}{c}{PASCAL-S}  \\
   & $\mathcal{S} \uparrow$ & $\mathcal{F}_{m} \mkern-7mu \uparrow$ & $\mathcal{E}_{\xi} \uparrow$ & $\mathcal{M} \downarrow$ 
   & $\mathcal{S} \uparrow$ & $\mathcal{F}_{m} \mkern-7mu \uparrow$ & $\mathcal{E}_{\xi} \uparrow$ & $\mathcal{M} \downarrow$
   & $\mathcal{S} \uparrow$ & $\mathcal{F}_{m} \mkern-7mu \uparrow$ & $\mathcal{E}_{\xi} \uparrow$ & $\mathcal{M} \downarrow$
    \\
    \midrule
    DUTS & .860 & .841 & .899 & .045 & .933 & .958 & .937 & .028 & .883 & .894 & .889 & .050 \\
    BDS  & \textbf{.875} & \textbf{.855} & \textbf{.910} & \textbf{.043} & \textbf{.935} & \textbf{.959} & \textbf{.941} & \textbf{.027} & \textbf{.891} & \textbf{.901} & \textbf{.909} & \textbf{.045} \\
    \bottomrule
    \bottomrule
  \end{tabular}
  \vspace{-0.1in}
\end{table}

\subsection{Influence of Edge Decoder. }
This experiment aims to explore the impact of our edge decoder on model performance, with the results shown in ~\cref{tab:decoder}. It is evident that without the dynamic edge-preserving decoder, the model performs poorly on more challenging datasets, indicating that it is only capable of aggregating simple features, leading to a decline in overall performance. In contrast, when utilizing the dynamic edge-preserving decoder, our model can progressively restore the resolution of the mask features while maintaining attention to the object edges, thereby improving the quality of the feature maps. This improvement demonstrates that the edge decoder helps retain fine-grained details and enhances the model's ability to handle complex scenarios, leading to a significant boost in performance across more difficult datasets.

 \begin{table}[!ht]
  \centering
  \caption{ Effectiveness of The Decoder.}
  \label{tab:decoder}
\renewcommand\tabcolsep{2.0pt}
  \renewcommand{\arraystretch}{1.33}
  \vspace{-0.1in}
  \begin{tabular}{c|cccc|cccc|cccc}
    \toprule
    \toprule
    \multirow{2}[1]{*}{} & \multicolumn{4}{c|}{DUT-O} & \multicolumn{4}{c|}{ECSSD}  & \multicolumn{4}{c}{PASCAL-S}  \\
   & $\mathcal{S} \uparrow$ & $\mathcal{F}_{m} \mkern-7mu \uparrow$ & $\mathcal{E}_{\xi} \uparrow$ & $\mathcal{M} \downarrow$ 
   & $\mathcal{S} \uparrow$ & $\mathcal{F}_{m} \mkern-7mu \uparrow$ & $\mathcal{E}_{\xi} \uparrow$ & $\mathcal{M} \downarrow$
   & $\mathcal{S} \uparrow$ & $\mathcal{F}_{m} \mkern-7mu \uparrow$ & $\mathcal{E}_{\xi} \uparrow$ & $\mathcal{M} \downarrow$
    \\
    \midrule
    w/o & .855 & .834 & .893 & .049 & \textbf{.933} & .957 & .936 & .028 & .881 & .891 & .884 & .052 \\
    w/  & \textbf{.860} & \textbf{.841} & \textbf{.899} & \textbf{.045} & \textbf{.933} & \textbf{.958} & \textbf{.937} & \textbf{.028} & \textbf{.883} & \textbf{.894} & \textbf{.889} & \textbf{.050} \\
    \bottomrule
    \bottomrule
  \end{tabular}
  \vspace{-0.1in}
\end{table}

\section{Conclusion}

In this work, our primary objective is to explore the transfer of knowledge from large multimodal models to generate accurate pseudo-labels, thereby reducing manual annotation costs. We introduce a novel dataset, BDS-TR, based on our framework for generating pseudo-labels. This dataset is significantly larger than previous datasets and encompasses a greater diversity of object categories and scenes, enabling our model to be applied across a broader range of scenarios.
We have demonstrated that the pseudo-labels generated through our approach exhibit high accuracy, which greatly enhances the model's overall performance. In terms of model architecture, we developed an edge-preserving decoder based on dynamic upsampling. This innovative decoder allows us to progressively increase the mask's resolution while preserving critical edge details.
Evaluation results indicate that our proposed method achieves state-of-the-art (SOTA) performance across all five benchmark datasets, underscoring the effectiveness and robustness of our approach. This work not only contributes to the field of SOD but also paves the way for future research in leveraging large models for improved visual tasks.

\newpage

\bibliographystyle{IEEEtran}
\bibliography{main}{}

\vfill

\end{document}